\documentclass[11pt]{article}

\usepackage[preprint]{acl}

\usepackage{times}
\usepackage{latexsym}
\usepackage{makecell}

\usepackage[T1]{fontenc}

\usepackage[utf8]{inputenc}

\usepackage{microtype}

\usepackage{inconsolata}

\usepackage{graphicx}

\usepackage{amsmath}
\usepackage{amssymb}
\usepackage{amsfonts}
\usepackage{booktabs}
\usepackage{graphicx}
\usepackage{tabularx}
\usepackage{hyperref}
\title{Beyond Uniform Forgetting: A Study of Sequential Direct Preference Optimization Across Preference Settings}

\author{
 \textbf{Pranav Bhandari\textsuperscript{1,2,*}}, 
  \textbf{Nicolas Fay\textsuperscript{3}},
  \textbf{Amitava Datta\textsuperscript{2}},
  \textbf{Usman Naseem\textsuperscript{4}},
 \textbf{Mehwish Nasim\textsuperscript{1,2,*}}
\\
 \textsuperscript{1}Network Analysis and Social Influence Modelling (NASIM) Lab \\
 \textsuperscript{2}School of Physics Maths and Computing, The University of Western Australia\\
 \textsuperscript{3}School of Psychological Science, The University of Western Australia \\
 \textsuperscript{4}School of Computing, Macquarie University
\\
  \small{
   \textbf{\textsuperscript{*}Correspondence:} {firstname.lastname@uwa.edu.au}
 }
}

\begin{document}
\maketitle
\begin{abstract}
Aligning language models with human preferences often requires optimising multiple behavioural objectives. A practical approach is to apply these objectives sequentially using preference optimisation methods such as Direct Preference Optimisation (DPO), but it remains unclear whether later training uniformly degrades preferences learned earlier or whether the effect depends on the relationship between objectives. We study sequential DPO across four preference settings covering distributional conflict, multi-attribute interaction, strong safety signal, and compatible response-quality objectives. Using Llama-3.1-8B-Instruct with LoRA adapters, we evaluate all objectives after every stage with a fixed base-model reference. We find that sequential DPO does not produce a single forgetting pattern; preference change ranges from partial degradation to stability, pair-level redistribution, or positive transfer depending on objective relationship, signal strength, and training order. Pair-level analysis using length-normalised policy margins shows that aggregate metrics can mask heterogeneous changes across preference pairs, whereas quartile decomposition reveals that high-confidence pairs can either degrade or improve depending on the setting. Mechanistic diagnostics show that Stage~2 gradients and adapter updates are near-orthogonal to the previous objective across all settings, providing little evidence that direct gradient opposition is the primary driver. These findings suggest that future sequential alignment pipelines should account for objective compatibility and signal strength, rather than assuming that later objectives affect earlier preferences uniformly.
\end{abstract}


\section{Introduction and Background}

Aligning large language models (LLMs) with human preferences plays a crucial role in their useful deployment and protection from negative social impacts \cite{safe_rlhf_pku, instructgpt, safe_rlhf_paper}. Multiple behavioural objectives, such as helpfulness, harmlessness, safety, honesty, factuality, style, and instruction following, are optimised in post-training pipelines \cite{safe_rlhf_pku, rlhf_paper, cui2023ultrafeedback, helpsteer2}. This usually involves optimising a policy against human feedback signals, either through reinforcement learning from human feedback (RLHF)~\cite{rlhf_paper} or through offline contrastive objectives such as Direct Preference Optimisation (DPO)~\cite{dpo_paper} and its variants such as IPO~\cite{ipo_paper} and SimPO ~\cite{simpo_paper,tie2025survey, xu-etal-2024-aligning}. These methods have driven significant progress in making models more helpful, less harmful, and better at following instructions. However, in practice, alignment is rarely a single-objective procedure \cite{lifelong_alignment}, and optimisation of multiple objectives is required \cite{zhong2024panacea, safe_rlhf_pku}. Post-training pipelines optimise for multiple behavioural dimensions, and preference data for different objectives are collected under different annotation protocols and task distributions~\cite{cui2023ultrafeedback,zhong2024panacea}. This creates a practical need to add new objectives without retraining from scratch.

Sequential direct preference optimisation can offer a practical way to incrementally extend alignment. Instead of retraining a model from scratch when new preference data becomes available, later objectives can be optimised from the previously aligned model, combining optimising methods like DPO-style preference optimisation with the sequential multi-dimensional alignment setting~\cite{sequential_preference_opt,dpo_paper}. This setup naturally raises a question about retention that when a later objective is optimised, to what extent are preferences learned in earlier stages preserved or degraded? Prior work on lifelong or sequential alignment has shown that adapting models to new preferences can lead to forgetting of earlier alignment behaviours~\cite{lifelong_alignment}, while recent large-scale studies of LLM post-training suggest a more nuanced picture in which forgetting is often moderate, task-dependent, and sensitive to the evaluation protocol~\cite{mapping_post_train_forgetting}.


Various evaluation methods are used to understand the effectiveness of alignment \cite{fu-etal-2025-minority, evaluating_alignment}. Aggregate task-level accuracy can hide substantial sample-level variation in what models retain or lose after post-training. Hence, introducing a sample-wise perspective is important because it moves beyond coarse average scores and provides a finer view of forgetting and backward transfer. However, existing analyses have largely focused on multiple-choice knowledge benchmarks, where forgetting is measured through correctness transitions and often relies on assumptions about random guessing or chance correction \cite{mapping_post_train_forgetting}. Such settings differ from preference alignment, where models are optimised over open-ended chosen--rejected response pairs and degradation may appear as changes in relative preference margins rather than discrete answer flips. Moreover, prior work primarily documents which post-training regimes produce forgetting or transfer, but provides limited evidence about \emph{why} these effects occur, especially whether they arise from direct objective conflict, data-distribution mismatch, differences in objective signal strength, or indirect parameter drift \cite{mapping_post_train_forgetting}.


In this work, we study preference degradation under sequential DPO. We investigate whether sequential DPO causes uniform forgetting of previous objectives, or whether the effect depends on the relationship between objectives and the individual preference pairs being evaluated. A natural candidate explanation for observed forgetting is direct gradient conflict between objectives. We test this directly by measuring gradient cosine similarity during sequential training, which allows us to distinguish active gradient opposition from more indirect mechanisms such as distributional drift or signal imbalance. To answer these questions, we evaluate models after every training stage on all objectives and analyse preference change at both aggregate and pair levels. For aggregate evaluation, we report DPO-relative reward margins and relative preference accuracy using a fixed base model as the evaluation reference. For pair-level diagnostics, we track length-normalised policy margins for individual chosen--rejected pairs across checkpoints.



Our experiments compare sequential DPO across four preference settings consisting of distributional conflict, multi-attribute interaction, strong safety signal, and compatible objectives. Across both objective orderings in each setting, we evaluate current-objective learning and previous-objective retention. Our results show that sequential DPO does not yield a uniform forgetting pattern; instead, preference change ranges from partial degradation to stability, redistribution, or positive transfer depending on objective relationship, signal strength, and training order.




 Pair-level analysis is conducted apart from aggregate trends which show that sequential DPO changes are heterogeneous i.e., some preference pairs lose margin while others improve, and the affected subset varies by dataset and ordering. Mechanistically, we find that Stage~2 updates are largely near-orthogonal to the previous objective, both in gradient space where measured and in LoRA \cite{lora_paper} adapter movement. This suggests that preference change is not primarily driven by direct gradient opposition, but is more consistent with drift, signal imbalance, or indirect parameter movement.

\subsection{Contributions}
\noindent\textbf{1.} We present a systematic empirical study of sequential alignment using DPO across four preference settings, showing that preference change spans a spectrum from bidirectional forgetting to positive transfer determined by objective relationship and signal strength rather than by sequential training alone.

\noindent\textbf{2.} We introduce pair-level policy-margin analysis for preference optimisation, tracking per-example margin changes across checkpoints using quartile decomposition, revealing heterogeneous redistribution that is hidden by aggregate reward margins and relative preference accuracy.

\noindent\textbf{3.} We provide mechanistic diagnostics via gradient conflict and adapter movement, showing that preference changes occur despite near-orthogonal update directions across all studied settings. This suggests indirect mechanisms rather than direct gradient opposition, although we interpret the evidence as suggestive because it is restricted to LoRA-parameter space.

\begin{figure*}[t]
\centering
\includegraphics[width=0.92\textwidth]{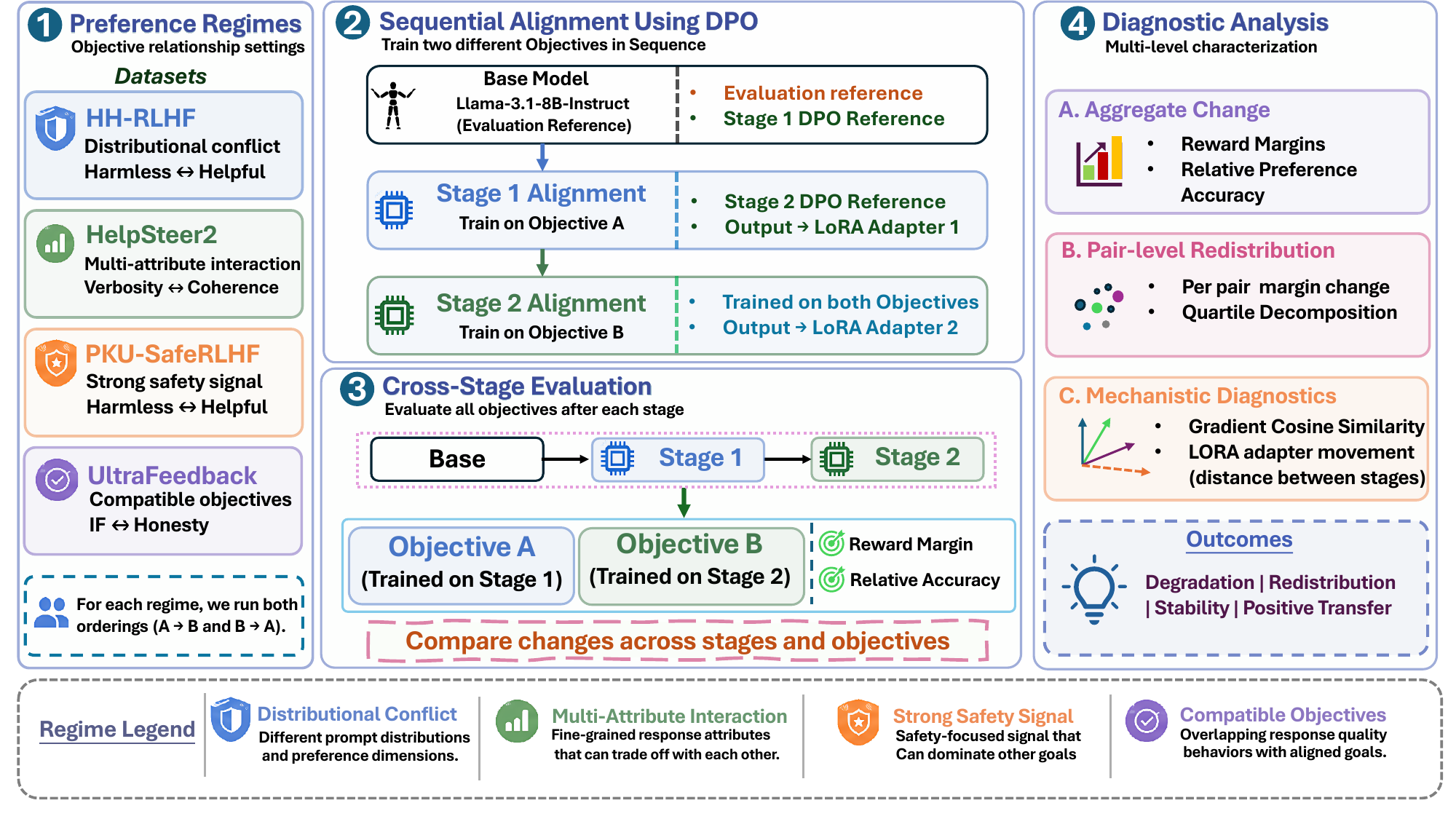}
\caption{
Overview of our sequential DPO evaluation pipeline. We compare four objective-relationship regimes, train objectives sequentially with LoRA adapters, evaluate all checkpoints against a fixed base reference, and analyse aggregate, pair-level, and mechanistic signatures of preference change.
}
\label{fig:methodology_overview}
\end{figure*}

\section{Experimental Setup}
\label{sec:setup}

\subsection{Overview}
\label{sec:setup:overview}

Preference-based alignment methods, including RLHF~\cite{rlhf_paper,instructgpt} and offline objectives such as DPO~\cite{dpo_paper} and IPO~\cite{ipo_paper}, rely on pairwise preference data. Each training example typically consists of a prompt $x$ and two responses: a preferred response $y_w$ and a rejected response $y_l$. We compare sequential DPO across four preference settings: \emph{distributional shift (HH-RLHF)}, \emph{multi-attribute interaction (HelpSteer2)}, \emph{strong signal safety (PKU-SafeRLHF)}, and \emph{compatible response-quality objectives (UltraFeedback)}. Across all experiments, a model is trained sequentially on two separate preference objectives using Direct Preference Optimisation (DPO). After each stage, we evaluate the resulting model on all objectives, including objectives trained in previous stages, allowing us to measure \textbf{preference degradation}, \textbf{transfer}, \textbf{stability}, and \textbf{pair-level changes} across training. Figure~\ref{fig:methodology_overview} summarises the overall experimental pipeline.

Our main experiments use Llama-3.1-8B-Instruct\footnote{\href{https://huggingface.co/meta-llama/Llama-3.1-8B-Instruct}{Llama-3.1-8B-Instruct}}~\cite{llama3_heard, llama_paper} as the base model with LoRA adapters. We use the instruction-tuned variant because it provides a realistic aligned starting point and a consistent conversational prompting format across preference datasets. Full model, LoRA, and training hyperparameters are reported in Appendix~\ref{app:implementation_details}.

\subsection{Sequential DPO Training}

We use Direct Preference Optimisation (DPO) \cite{dpo_paper} to train on each preference objective. Given a prompt $x$ and a pair of responses $(y_w, y_l)$, where $y_w$ is the preferred response and $y_l$ is the rejected response, DPO optimises the policy $\pi_\theta$ relative to a frozen reference policy $\pi_{\mathrm{ref}}$:

\begin{equation}
\small
h_\theta(x,y_w,y_l)
=
\log \frac{\pi_\theta(y_w|x)}{\pi_\theta(y_l|x)}
-
\log \frac{\pi_{\mathrm{ref}}(y_w|x)}{\pi_{\mathrm{ref}}(y_l|x)},
\label{eq:dpo_gap}
\end{equation}
\begin{equation}
\small
\mathcal{L}_{\mathrm{DPO}}
=
-\mathbb{E}_{(x,y_w,y_l)\sim\mathcal{D}}
\left[
\log \sigma
\left(
\beta\, h_\theta(x,y_w,y_l)
\right)
\right]
\label{eq:dpo}
\end{equation}
where $h_\theta(x,y_w,y_l)$ is the relative log-ratio between policy and reference.

In sequential training, each stage resumes from the adapter checkpoint produced by the previous stage. The DPO reference model for each stage is the model state at the beginning of that stage. For example, the Stage~2 reference is the final Stage~1 adapter. \emph{This mirrors practical sequential fine-tuning pipelines, where each new alignment stage starts from the previously aligned model.} This shifting reference applies only during training. For evaluation, we always use the fixed base model as the reference, ensuring that reward margins are on a consistent scale across all stages and checkpoints.

\subsection{Preference-Dataset Taxonomy}
\label{sec:dataset_taxonomy}

A central goal of this work is to test whether sequential DPO behaves differently under different relationships between preference objectives. We therefore compare four settings, each containing two objectives and both training orderings.

\paragraph{Distributional conflict: HH-RLHF.}
We use the \emph{helpful-base} and \emph{harmless-base} subsets of HH-RLHF~\cite{rlhf_paper} to study a safety--helpfulness shift. These objectives differ in both prompt distribution and annotated preference dimension, but our diagnostics find no flipped-label pairs and less than $1\%$ prompt overlap ($0.86\%$), suggesting a primarily distributional rather than adversarial conflict.

\paragraph{Multi-attribute interaction: HelpSteer2.}
 HelpSteer2~\cite{helpsteer2} is used to construct preference objectives for verbosity and coherence. This setting tests interaction between fine-grained response attributes, where increased detail can interact with logical organisation and focus.

\paragraph{Strong safety signal: PKU-SafeRLHF.}
We use PKU-SafeRLHF~\cite{safe_rlhf_pku} to study a safety--helpfulness setting in which harmlessness provides a much stronger training signal than in HH-RLHF. This tests whether a high-signal safety objective is more resistant to later-stage preference change.

\paragraph{Compatible objectives: UltraFeedback.}
We use UltraFeedback~\cite{cui2023ultrafeedback} to construct instruction-following and honesty objectives. These objectives reward overlapping response-quality behaviours, providing a compatible-objective baseline.

Together, these settings cover a spectrum of preference-relationship regimes under a shared sequential DPO and cross-evaluation protocol. Further dataset construction details are provided in Appendix~\ref{app:dataset_details}.

\section{Evaluation Protocol}

\paragraph{DPO-Relative Reward Margin.}
For a preference pair $(x,y_w,y_l)$, we report the DPO-relative reward margin \cite{dpo_paper}:
\begin{equation}
\small
r_\theta(x,y_w,y_l)
=
\beta
\left[
\Delta_\theta(x,y_w,y_l)
-
\Delta_{\mathrm{ref}}(x,y_w,y_l)
\right]
\label{eq:reward_margin}
\end{equation}
where,
\begin{equation}
\small
\Delta_\theta(x,y_w,y_l)
=
\log \pi_\theta(y_w|x)
-
\log \pi_\theta(y_l|x),
\label{eq:policy_logratio}
\end{equation}
\begin{equation}
\small
\Delta_{\mathrm{ref}}(x,y_w,y_l)
=
\log \pi_{\mathrm{ref}}(y_w|x)
-
\log \pi_{\mathrm{ref}}(y_l|x)
\label{eq:reference_logratio}
\end{equation}
where $\log \pi_\theta(y|x)$ denotes the sequence log-probability of response $y$ given prompt $x$.
Here, $\pi_\theta$ is the evaluated policy and $\pi_{\mathrm{ref}}$ is the frozen reference model. A positive reward margin means that the policy increases the chosen-over-rejected preference gap relative to the reference model. A value of zero means no relative change from the reference. We report the mean reward margin over all evaluation pairs.

\paragraph{Relative Preference Accuracy.}
We also report relative preference accuracy:
\begin{equation}
\small
\mathrm{Acc}_{\mathrm{rel}}
=
\frac{1}{|\mathcal{D}|}
\sum_{i=1}^{|\mathcal{D}|}
\left(
\mathbf{1}[r_i > 0]
+
\tfrac{1}{2}\,\mathbf{1}[r_i = 0]
\right)
\label{eq:relative_pref_accuracy}
\end{equation}
where $r_i = r_\theta(x_i,y_{w,i},y_{l,i})$. This metric measures the fraction of pairs for which the policy improves the chosen-over-rejected gap relative to the reference model, with ties assigned half credit. We report DPO-relative reward margin and relative preference accuracy, both computed using the fixed base model as the evaluation reference. This differs from the training DPO reference, which is the previous-stage adapter; keeping the evaluation reference fixed at the base model ensures that cross-stage margin comparisons are on a consistent scale. Thus, a zero reward margin indicates that there is no preference gap relative to the base model, and a relative preference accuracy of 0.5 corresponds to the neutral baseline.


\paragraph{Pairwise Forgetting Analysis.}
In addition to aggregate DPO-relative metrics, we analyse changes at the level of individual preference pairs. For this analysis, we compute a length-normalised policy preference margin for each pair:
\begin{equation}
\small
m_i
=
\overline{\log \pi_\theta}(y_{w,i}|x_i)
-
\overline{\log \pi_\theta}(y_{l,i}|x_i)
\label{eq:policy_margin}
\end{equation}
where $\overline{\log \pi_\theta}$ denotes the mean per-token log-probability over response tokens. This pairwise margin directly measures whether the policy itself assigns higher average token probability to the chosen response than to the rejected response.

For each pair $i$, we compute its margin at each checkpoint and define the per-pair margin change between Stage~1 and Stage~2 as:
\begin{equation}
\small
\Delta_i = m^{(1)}_i - m^{(2)}_i 
\label{eq:pairwise_drop}
\end{equation}

Positive values indicate that the pair's policy preference margin decreased after Stage~2, while negative values indicate that the margin increased. These changes are analysed using quartile decomposition.


\paragraph{Quartile Decomposition.}
For interpretability, we also sort evaluation pairs by their Stage~1 policy margin and divide them into four equally sized groups. Q1 contains the lowest-margin pairs, which we refer to as hard or low-confidence pairs. Q4 contains the highest-margin pairs, which we refer to as easy or high-confidence pairs. We then compute the mean margin change within each quartile:
\begin{equation}
\overline{\Delta}_{Q_k}
=
\frac{1}{|Q_k|}
\sum_{i\in Q_k}
\Delta_i 
\label{eq:quartile_drop}
\end{equation}
This shows whether the aggregate change is driven by hard pairs, easy pairs, or distributed uniformly across the evaluation set.

\subsection{Mechanistic Diagnostics}
\label{sec:setup:mechanistic}

\paragraph{Gradient Conflict.}
To test whether observed degradation is caused by direct opposition between objectives, we measure the cosine similarity between gradients during Stage~2 training. At measurement steps, we compare the live Stage~2 DPO gradient with a diagnostic Stage~1-objective gradient computed on a fixed reference batch. Gradients are collected only over trainable LoRA parameters. Cosine values near $-1$ indicate opposing objectives, values near $0$ indicate orthogonality, and values near $+1$ indicate alignment.


\paragraph{Adapter Movement.}
We additionally analyse inter-stage LoRA adapter movement as an auxiliary parameter-space diagnostic. We compute the raw LoRA parameter displacement between stage checkpoints(Appendix~\ref{app:adapter_movement}) and measure whether later-stage updates are aligned with, orthogonal to, or opposed to previous-stage adapter directions. Because the raw LoRA parameter space is not invariant to LoRA rescaling, this analysis is used as supporting evidence rather than as the primary mechanism test.

\section{Results}
\label{sec:results}

We organise the results around four questions. 

\noindent\textbf{1.} First, does sequential DPO produce measurable preference change, including degradation or transfer, across objectives?

\noindent\textbf{2.} Second, are these changes uniform across examples, or concentrated in specific subsets of preference pairs? 

\noindent\textbf{3.} Third, does the relationship between objectives determine whether sequential training produces forgetting, redistribution, stability, or positive transfer? 

\noindent\textbf{4.} Finally, can observed preference changes be explained by direct gradient-level conflict, or do they point to drift or indirect parameter movement?

Across all experiments, evaluation is deterministic, i.e., we score fixed chosen/rejected responses using model log-probabilities rather than model generations. For pair-level forgetting diagnostics, we use length-normalised policy margins (Eq.~\ref{eq:policy_margin}), which measure the policy's absolute chosen-over-rejected log-probability preference independent of any reference model. The held-out evaluation pairs are used for pair-level analysis. 

\subsection{Sequential DPO Produces Measurable Preference Change, Not Uniform Forgetting}

We first examine aggregate cross-stage performance. Table~\ref{tab:aggregate_results} reports reward margin and relative preference accuracy after each training stage, evaluated on all objectives rather than only the current-stage objective. This allows us to distinguish direct learning of the current objective from degradation of previously trained objectives.

\begin{table*}[t]
\centering

\newcommand{\datasetsep}{
\specialrule{0.7pt}{1.5pt}{0.2pt}
\specialrule{0.25pt}{0pt}{1.2pt}
}

{
\scriptsize
\setlength{\tabcolsep}{3.0pt}
\renewcommand{\arraystretch}{0.95}

\begin{minipage}[t]{0.493\textwidth}
\centering

\begin{tabular}{llrrrr}
\toprule
\textbf{Stage} & \textbf{Eval Objective} & \textbf{Margin} & \textbf{$\Delta$ Margin} & \textbf{Accuracy} & \textbf{$\Delta$ Acc.} \\
\midrule
\multicolumn{6}{@{}l}{\textbf{HH-RLHF: harmless$\rightarrow$helpful}} \\
Base    & Harmless & 0.000  & --        & 50.0 & --       \\
Stage 1 & Harmless & 0.925  & $+$0.925  & 67.0 & $+$17.0 \\
Stage 1 & Helpful  & $-$0.508 & $-$0.508 & 38.1 & $-$11.9 \\
Stage 2 & Harmless & 0.707  & $-$0.218  & 61.4 & $-$5.6  \\
Stage 2 & Helpful  & 0.283  & $+$0.791  & 57.8 & $+$19.7 \\
\datasetsep
\multicolumn{6}{@{}l}{\textbf{HH-RLHF: helpful$\rightarrow$harmless}} \\
Base    & Helpful  & 0.000  & --        & 50.0 & --       \\
Stage 1 & Helpful  & 0.731  & $+$0.731  & 66.1 & $+$16.1 \\
Stage 1 & Harmless & $-$0.362 & $-$0.362 & 41.1 & $-$8.9  \\
Stage 2 & Helpful  & 0.228  & $-$0.503  & 55.4 & $-$10.7 \\
Stage 2 & Harmless & 0.609  & $+$0.971  & 59.7 & $+$18.6 \\
\datasetsep
\multicolumn{6}{@{}l}{\textbf{HelpSteer2: verbosity$\rightarrow$coherence}} \\
Base    & Verbosity & 0.000  & --        & 50.0 & --       \\
Stage 1 & Verbosity & 4.048  & $+$4.048  & 86.1 & $+$36.1 \\
Stage 1 & Coherence & $-$0.005 & $-$0.005 & 51.2 & $+$1.2  \\
Stage 2 & Verbosity & 3.776  & $-$0.273  & 80.1 & $-$6.0  \\
Stage 2 & Coherence & 1.023  & $+$1.028  & 59.8 & $+$8.6  \\
\datasetsep
\multicolumn{6}{@{}l}{\textbf{HelpSteer2: coherence$\rightarrow$verbosity}} \\
Base    & Coherence & 0.000  & --        & 50.0 & --       \\
Stage 1 & Coherence & 0.925  & $+$0.925  & 66.0 & $+$16.0 \\
Stage 1 & Verbosity & $-$0.363 & $-$0.363 & 45.3 & $-$4.7  \\
Stage 2 & Coherence & 0.720  & $-$0.205  & 57.0 & $-$9.0  \\
Stage 2 & Verbosity & 3.780  & $+$4.143  & 83.9 & $+$38.6 \\
\bottomrule
\end{tabular}
\end{minipage}
\hspace{0.006\textwidth}
\begin{minipage}[t]{0.493\textwidth}
\centering

\begin{tabular}{llrrrr}
\toprule
\textbf{Stage} & \textbf{Eval Objective} & \textbf{Margin} & \textbf{$\Delta$ Margin} & \textbf{Accuracy} & \textbf{$\Delta$ Acc.} \\
\midrule
\multicolumn{6}{@{}l}{\textbf{PKU-SafeRLHF: harmless$\rightarrow$helpful}} \\
Base    & Harmless & 0.000  & --        & 50.0 & --       \\
Stage 1 & Harmless & 14.071 & $+$14.071 & 95.6 & $+$45.6 \\
Stage 1 & Helpful  & 0.761  & $+$0.761  & 51.4 & $+$1.4  \\
Stage 2 & Harmless & 16.930 & $+$2.859  & 95.9 & $+$0.3  \\
Stage 2 & Helpful  & 1.575  & $+$0.814  & 58.7 & $+$7.3  \\
\datasetsep
\multicolumn{6}{@{}l}{\textbf{PKU-SafeRLHF: helpful$\rightarrow$harmless}} \\
Base    & Helpful  & 0.000  & --        & 50.0 & --       \\
Stage 1 & Helpful  & 0.622  & $+$0.622  & 67.3 & $+$17.3 \\
Stage 1 & Harmless & 1.318  & $+$1.318  & 73.5 & $+$23.5 \\
Stage 2 & Helpful  & 1.230  & $+$0.608  & 57.4 & $-$9.9  \\
Stage 2 & Harmless & 13.184 & $+$11.866 & 95.4 & $+$21.9 \\
\datasetsep
\multicolumn{6}{@{}l}{\textbf{UltraFeedback: IF$\rightarrow$honesty}} \\
Base    & IF      & 0.000  & --        & 50.0 & --       \\
Stage 1 & IF      & 8.195  & $+$8.195  & 85.1 & $+$35.1 \\
Stage 1 & Honesty & 7.471  & $+$7.471  & 84.3 & $+$34.3 \\
Stage 2 & IF      & 15.114 & $+$6.919  & 87.8 & $+$2.7  \\
Stage 2 & Honesty & 13.572 & $+$6.101  & 86.0 & $+$1.7  \\
\datasetsep
\multicolumn{6}{@{}l}{\textbf{UltraFeedback: honesty$\rightarrow$IF}} \\
Base    & Honesty & 0.000  & --        & 50.0 & --       \\
Stage 1 & Honesty & 5.331  & $+$5.331  & 82.4 & $+$32.4 \\
Stage 1 & IF      & 6.152  & $+$6.152  & 85.8 & $+$35.8 \\
Stage 2 & Honesty & 13.114 & $+$7.783  & 84.1 & $+$1.7  \\
Stage 2 & IF      & 15.745 & $+$9.593  & 87.9 & $+$2.1  \\
\bottomrule
\end{tabular}
\end{minipage}

}
\caption{
Aggregate DPO reward margin and relative preference accuracy computed with the fixed base model as evaluation reference.
Stage~1 rows report evaluation on both the trained objective and the untrained objective, showing cross-objective effects before Stage~2 begins.
$\Delta$ columns show change from the same objective's most recent checkpoint; negative values indicate degradation.
Accuracy values are percentages; $\Delta$ accuracy values are percentage-point changes.
}
\label{tab:aggregate_results}
\end{table*}

Our aggregate results show that sequential DPO changes both trained and cross-objective preferences, but the pattern does not support uniform catastrophic forgetting. Instead, later-stage training produces partial degradation, stability, or positive transfer at the aggregate level, with pair-level redistribution examined below. In HH-RLHF, later-stage training measurably degrades the previous objective while acquiring the new one. In the harmless$\rightarrow$helpful ordering, harmless reward margin drops from 0.925 to 0.707 after helpful training ($-$0.218, $-$23.6\%), while helpful accuracy recovers from 38.1\% after Stage~1 harmless training to 57.8\% after Stage~2 helpful training. In the helpful$\rightarrow$harmless ordering, helpful margin drops substantially from 0.731 to 0.228 ($-$0.503, $-$68.8\%), while harmless accuracy recovers from 41.1\% after Stage~1 helpful training to 59.7\% after Stage~2 harmless training. In both cases, the previous objective remains above the neutral baseline after Stage~2 (61.4\% and 55.4\%, respectively), indicating partial rather than catastrophic forgetting.

In HelpSteer2, where the attribute pair is constructed from verbosity and coherence, the pattern differs from HH-RLHF. In the verbosity$\rightarrow$coherence ordering, verbosity margin drops from 4.048 to 3.776 ($-$0.273, $-$6.7\%), while coherence reaches 1.023 margin and 59.8\% accuracy after Stage~2. In the coherence$\rightarrow$verbosity ordering, coherence margin drops from 0.925 to 0.720 ($-$0.205, $-$22.2\%), while verbosity reaches 3.780 margin and 83.9\% accuracy. Verbosity produces a much larger observed margin than coherence, which affects how degradation is interpreted in this setting.

In PKU-SafeRLHF, the harmless objective produces very large margins (Stage~1: 14.071, compared to 0.925 in HH-RLHF), reflecting the much stronger harmlessness signal in this construction. Consequently, harmless preferences do not degrade after helpful training: margin increases from 14.071 to 16.930 and accuracy remains at 95.9\%. In the helpful$\rightarrow$harmless ordering, helpful accuracy drops from 67.3\% to 57.4\% ($-$9.9pp) after harmless training, even though the DPO-relative helpful margin increases from 0.622 to 1.230. In this setting, the accuracy drop provides clearer evidence of helpfulness degradation because the margin and accuracy move in opposite directions.

In UltraFeedback, neither ordering shows aggregate forgetting of the previously trained objective. In IF$\rightarrow$honesty, IF margin increases from 8.195 to 15.114, and honesty reaches 13.572 after Stage~2. In honesty$\rightarrow$IF, honesty margin increases from 5.331 to 13.114, and IF reaches 15.745 after Stage~2. Moreover, after Stage~1 IF training alone, the honesty margin is already 7.471 with 84.3\% accuracy, and after Stage~1 honesty training alone, IF reaches 6.152 with 85.8\% accuracy. This suggests substantial compatibility between instruction-following and honesty in this construction, providing a complementary baseline where sequential DPO produces positive transfer rather than interference.

\subsection{Pair-Level Analysis Reveals Heterogeneous Preference Redistribution}

We next move from aggregate evaluation to pair-level analysis. For each held-out preference pair $i$, we compute its margins at Stages~1 and~2 and define the pair-level margin change ($\Delta_i$) as given in Equation \ref{eq:pairwise_drop}. Positive $\Delta_i$ indicates that the preference margin decreased after the next training stage; negative $\Delta_i$ indicates that the margin increased.

\begin{table}[t]
\centering
\scriptsize
\setlength{\tabcolsep}{2.4pt}
\renewcommand{\arraystretch}{0.95}
\resizebox{\columnwidth}{!}{
\begin{tabular}{llcrr}
\toprule
Dataset/Order & Obj. & $n$ & Mean $\Delta$ & Improved \\
\midrule
HH Ha$\rightarrow$He & Harmless & 1{,}000 & $+$0.013 & 47.5\% \\
HH He$\rightarrow$Ha & Helpful  & 1{,}000 & $+$0.042 & 39.3\% \\
HS2 Ve$\rightarrow$Co & Verbosity & 527 & $-$0.023 & 60.9\% \\
HS2 Co$\rightarrow$Ve & Coherence & 422 & $-$0.004 & 43.1\% \\
PKU Ha$\rightarrow$He & Harmless & 1{,}000 & $-$0.080 & 71.8\% \\
PKU He$\rightarrow$Ha & Helpful  & 1{,}000 & $-$0.024 & 51.3\% \\
UF IF$\rightarrow$Hon & IF       & 699     & $-$0.314 & 80.8\% \\
UF Hon$\rightarrow$IF & Honesty  & 697     & $-$0.332 & 78.3\% \\
\bottomrule
\end{tabular}
}
\caption{
Pair-level margin change after sequential training. Mean $\Delta$ is the average change in length-normalised policy margin, where positive values indicate degradation and negative values indicate improvement. Improved reports the fraction of evaluation pairs whose margin increased after Stage~2.
}
\label{tab:pairwise_change}
\end{table}
We use Stage~1 margin as an approximate baseline for model confidence, i.e., high-margin pairs are ``easy'' because the model strongly prefers the chosen response, while low-margin pairs are ``hard'' because the preference is weaker or more uncertain. Table~\ref{tab:pairwise_change} reports pair-level change statistics across all eight orderings. Mean change alone is insufficient because it can hide substantial heterogeneity. Thus, a model may preserve aggregate accuracy while still changing a meaningful subset of preference pairs.

For HH-RLHF, the two orderings show different pair-level behaviour. In harmless$\rightarrow$helpful, the mean change is small ($+0.013$) and the improved fraction is close to balanced (47.5\%), suggesting only mild pair-level shift. In helpful$\rightarrow$harmless, the mean change is larger and positive ($+0.042$), while only 39.3\% of pairs improve, indicating broader helpfulness degradation after harmlessness training. As shown in Figure~\ref{fig:quartile_results}, this degradation is concentrated most strongly among easy/high-confidence helpfulness pairs.

Overall, the pair-level results show that sequential DPO does not simply raise or lower all preferences uniformly. Later training can degrade some previously learned preferences while preserving or improving others, producing structured preference change across the evaluation set. The following quartile analysis examines where these changes concentrate across the Stage~1 confidence distribution.

\begin{table*}[t]
\centering
\small
\resizebox{\textwidth}{!}{
\begin{tabular}{llcccccc}
\toprule
Dataset & Order & Grad.\ cosine & Grad.\ angle & $\|S_1\|$ & Rel.\ move & Adapter dir.\ cos. & Verdict \\
\midrule
HH-RLHF & harm.$\rightarrow$help. & $-$0.004 & $90.2^\circ$ & 48.86 & 9.19\% & $-$0.007 ($90.4^\circ$) & near-orthogonal \\
HH-RLHF & help.$\rightarrow$harm. & $-$0.007 & $90.4^\circ$ & 48.85 & 9.14\% & $-$0.007 ($90.4^\circ$) & near-orthogonal \\
HelpSteer2 & verb.$\rightarrow$cohere. & $+$0.000 & $90.0^\circ$ & 48.79 & 4.54\% & $-$0.002 ($90.1^\circ$) & near-orthogonal \\
HelpSteer2 & cohere.$\rightarrow$verb. & $-$0.002 & $90.1^\circ$ & 48.77 & 5.12\% & $-$0.002 ($90.1^\circ$) & near-orthogonal \\
PKU-SafeRLHF & harm.$\rightarrow$help. & $-$0.003 & $90.1^\circ$ & 48.94 & 9.31\% & $-$0.001 ($90.1^\circ$) & near-orthogonal \\
PKU-SafeRLHF & help.$\rightarrow$harm. & $+$0.028 & $88.4^\circ$ & 48.88 & 9.09\% & $+$0.004 ($89.8^\circ$) & near-orthogonal \\
UltraFeedback & IF$\rightarrow$hon. & $+$0.004 & $89.7^\circ$ & 49.44 & 16.12\% & $+$0.002 ($89.9^\circ$) & near-orthogonal \\
UltraFeedback & hon.$\rightarrow$IF & $-$0.001 & $90.1^\circ$ & 49.43 & 16.02\% & $+$0.001 ($89.9^\circ$) & near-orthogonal \\
\bottomrule
\end{tabular}
}
\caption{
Mechanistic diagnostics across all four datasets and orderings ($n{=}5$--8 logging steps per run). Gradient cosine compares the live Stage~2 DPO gradient against a fixed previous-objective reference gradient over LoRA parameters. Relative move is computed as $\|S_2-S_1\|/\|S_1\|$ using flattened raw LoRA adapter parameters; adapter direction cosine is $\cos(S_1,S_2-S_1)$.
}
\label{tab:mechanistic_diagnostics}
\end{table*}
\subsection{Quartile Decomposition Shows Where Change Concentrates}
\label{sec:results:quartiles}

\begin{figure}[t]
\centering
\includegraphics[width=\columnwidth]{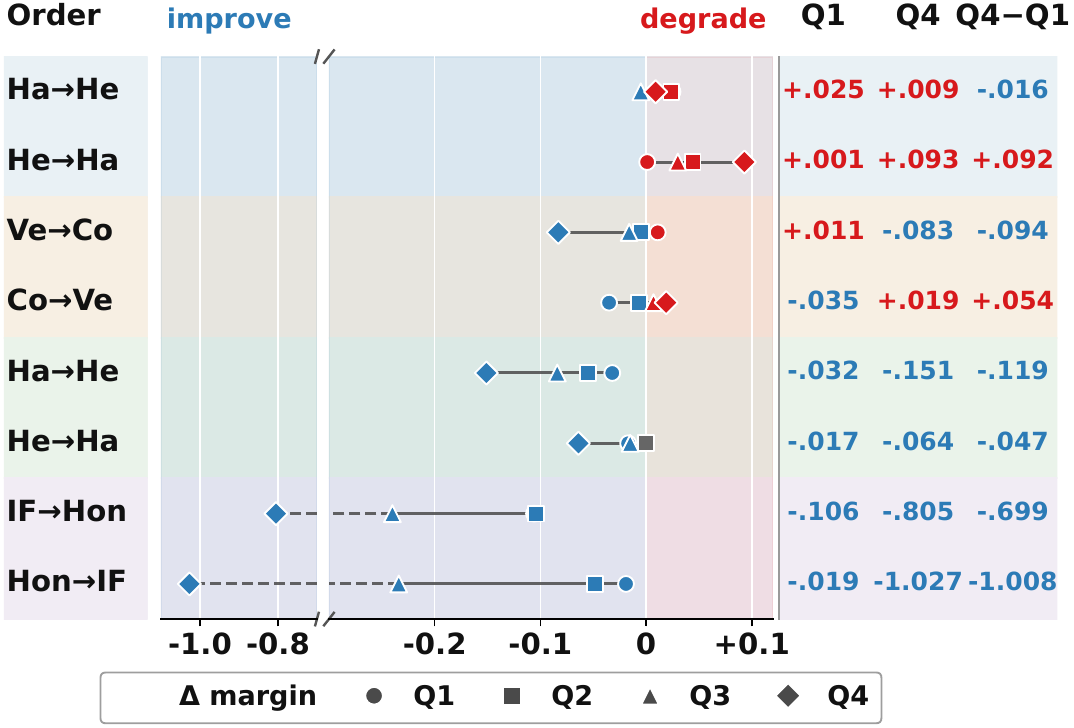}
\caption{
Quartile decomposition of pair-level margin change after sequential training. 
The figure shows the quartile profile for each ordering and reports Q1, Q4, and Q4--Q1 to emphasise where change concentrates. 
Blue indicates improvement ($\Delta<0$), while red indicates degradation ($\Delta>0$). 
Row strips correspond, from top to bottom, to HH-RLHF, HelpSteer2, PKU-SafeRLHF, and UltraFeedback.
}
\label{fig:quartile_results}
\end{figure}

To identify where preference change concentrates, we sort evaluation pairs by their Stage~1 margin and divide them into quartiles. Q1 contains the hardest/lowest-margin pairs, while Q4 contains the easiest/highest-margin pairs. Figure~\ref{fig:quartile_results} summarises the hardest--easiest contrast using Q1, Q4, and Q4--Q1; full Q1--Q4 values are reported in Appendix Table~\ref{tab_app:quartile_results}. Negative values indicate margin improvement and positive values indicate degradation.

In PKU-SafeRLHF (harmless$\rightarrow$helpful ordering), all four quartiles 
of harmless pairs improve after helpful training: Q1 $-$0.032, Q2 $-$0.055,
Q3 $-$0.084, and Q4 $-$0.151. The improvement grows monotonically from hard 
to easy pairs, showing that high-confidence harmless pairs are reinforced
rather than degraded. This is directionally opposite to the HH-RLHF        
helpful$\rightarrow$harmless pattern, where Q4 shows the largest
degradation. In the helpful$\rightarrow$harmless ordering, changes are
smaller and less structured across quartiles, consistent with the weaker
helpfulness signal in PKU-SafeRLHF.

In UltraFeedback, improvement is concentrated in the highest-confidence     
pairs for both orderings. In IF$\rightarrow$honesty, Q1 and Q2 show modest
improvement ($-$0.106 and $-$0.104), while Q3 and Q4 show substantially     
larger gains ($-$0.240 and $-$0.805). In honesty$\rightarrow$IF, the pattern
is similar: Q1 $-$0.019, Q2 $-$0.048, Q3 $-$0.234, and Q4 $-$1.027. The
highest-confidence pairs benefit most from the second stage. This stands in
direct contrast to the HH-RLHF helpful$\rightarrow$harmless case, where the
highest-confidence pairs lose the most margin. In both cases, the largest
changes concentrate in Q4, but the direction is reversed.

Overall, the quartile analysis shows that sequential DPO does not uniformly 
raise or lower an objective across all pairs. Later training can act
selectively across the confidence distribution: in some settings,
high-confidence pairs are most degraded, while in others they are most
reinforced. This suggests that where change concentrates depends on the
relationship between the two objectives, rather than following a single
uniform forgetting pattern.
\subsection{Mechanistic Diagnostics: Gradient Conflict and Adapter Movement}
\label{sec:results:mechanistic}

To test whether observed preference changes are caused by direct objective opposition, we examine two complementary diagnostics: \textbf{gradient cosine similarity during Stage~2 training} and \textbf{raw LoRA adapter movement between stage checkpoints}. The results for all four datasets and orderings are reported in Table~\ref{tab:mechanistic_diagnostics}.

\paragraph{Gradient conflict:}
Gradient cosine similarity is near zero across all eight orderings, with mean cosine ranging from $-0.007$ to $+0.028$ and mean angle ranging from $88.4^\circ$ to $90.4^\circ$. This indicates that Stage~2 DPO gradients are near-orthogonal to previous-objective reference gradients in LoRA parameter space. No ordering shows evidence of systematic gradient opposition across the measured steps. The largest deviation from zero occurs in PKU-SafeRLHF helpful$\rightarrow$harmless ($+0.028$, $88.4^\circ$), indicating slight alignment rather than opposition. This is consistent with the same run where the DPO-relative helpful margin increases after harmless training (Table~\ref{tab:aggregate_results}). All other orderings fall within $|\cos| \leq 0.007$, effectively near $90^\circ$.

\paragraph{Adapter movement:}
Stage~2 training moves only 4.5--16.1\% of the Stage~1 adapter norm across experiments, while the update direction is near-orthogonal to the Stage~1 adapter in every case ($\cos \approx 0$, $\theta \approx 90^\circ$). HH-RLHF moves 9.1--9.2\%, HelpSteer2 moves 4.5--5.1\%, PKU-SafeRLHF moves 9.1--9.3\%, and UltraFeedback moves 16.0--16.1\%. Thus, even when the update magnitude varies across datasets, later training does not directly reverse the previous-stage adapter; it moves sideways in raw LoRA parameter space.

Together, these diagnostics separate direct opposition from indirect movement. Strongly negative gradients or adapter directions would suggest that later training actively pushes against the previous objective. Instead, we observe near-zero gradient cosine and near-orthogonal adapter movement across all settings. These results are more consistent with distributional drift, signal imbalance, indirect representational movement, or loss of exposure to previous-objective data than with direct gradient opposition. We interpret this evidence as suggestive rather than causal, because gradient measurements are limited to LoRA parameters and a fixed reference batch, and raw LoRA parameter space is not invariant to rescaling.

\section{Conclusion}

We studied whether sequential DPO uniformly degrades earlier preference objectives. Across four preference settings, we find that it does not; later training can lead to partial degradation, stability, pair-level redistribution, or positive transfer depending on the relationship between objectives, their signal strength, and training order. Our results also show that aggregate metrics alone can hide where these changes occur, motivating pair-level and quartile analyses for understanding sequential preference change. Mechanistic diagnostics provide little evidence that direct gradient opposition is the main explanation, suggesting instead that objective relationship and signal strength play an important role. Future work should extend these diagnostics to larger models, additional alignment objectives, and methods that explicitly account for objective compatibility when designing sequential post-training pipelines.

\section*{Limitations}

Our experiments are conducted with Llama-3.1-8B-Instruct using LoRA adapters, so future work should test whether the findings generalise to other model families, model scales, and full-parameter fine-tuning. We study four preference-relationship regimes, but additional datasets and objectives such as factuality, reasoning, refusal style, and domain-specific safety would provide a broader view of sequential alignment. We also use consistent hyperparameters across objectives to keep comparisons controlled and interpretable; objective-specific schedules or adaptive stopping may further improve individual stages. Finally, our mechanistic diagnostics focus on LoRA gradients and adapter movement, which makes the analysis tractable for 8B-scale experiments but does not capture all full-model or activation-level dynamics.

\section*{Ethical Considerations}

This work analyses how sequential preference training can change or degrade previously learned alignment behaviours. We use publicly available preference datasets and evaluate fixed chosen--rejected responses using log-probabilities rather than generating new outputs, reducing the risk of producing harmful content during evaluation. Some datasets contain safety-related prompts and harmful rejected responses, which are used only for diagnostic evaluation. The intended use of this work is to improve monitoring of sequential alignment pipelines, not to weaken safety behaviours or intentionally induce preference degradation.

\bibliography{custom}

\appendix

\newpage

\section{Model, Training, and Implementation Details}
\label{app:implementation_details}

\subsection{Base Model}
\label{app:model}

All experiments use Llama-3.1-8B-Instruct \cite{llama3_heard} as the base language model. The instruction-tuned model is used because it represents a realistic starting point for sequential alignment and provides a consistent chat template for formatting preference data. The model contains approximately 8.1B parameters. All experiments are run using bfloat16 precision with gradient checkpointing enabled on a single NVIDIA GPU with 48GB memory.

\subsection{LoRA Configuration}
\label{app:lora}

We use Low-Rank Adaptation (LoRA) for all fine-tuning runs. The pretrained base model weights are frozen, and only LoRA adapter parameters are updated. We apply LoRA to the following projection matrices in every transformer layer:

\begin{itemize}
    \item query projection: \texttt{q\_proj}
    \item key projection: \texttt{k\_proj}
    \item value projection: \texttt{v\_proj}
    \item output projection: \texttt{o\_proj}
    \item up projection: \texttt{up\_proj}
    \item down projection: \texttt{down\_proj}
    \item gate projection: \texttt{gate\_proj}
\end{itemize}

Unless otherwise stated, we use LoRA rank $r=32$, scaling factor $\alpha=128$, dropout $p=0.05$, and no bias adaptation. This gives 83,886,080 trainable parameters, corresponding to approximately 1.03\% of the full model parameters.

LoRA is useful for this study because it enables exact adapter checkpointing after each sequential stage. This makes it possible to compare model states across stages without storing full model checkpoints.

\subsection{DPO Training Hyperparameters}
\label{app:dpo_hparams}

All training uses the HuggingFace TRL \texttt{DPOTrainer} with the sigmoid DPO loss. Unless otherwise stated, we use the following hyperparameters:

\begin{itemize}
    \item learning rate: $5\times10^{-5}$
    \item learning rate schedule: cosine
    \item warmup ratio: 10\% of total optimisation steps
    \item maximum gradient norm: 1.0
    \item maximum total sequence length: 1024 tokens
    \item maximum prompt length: 512 tokens
    \item precision: bfloat16
    \item gradient checkpointing: enabled
    \item random seed: 42
\end{itemize}

For HH-RLHF experiments, we use a per-device batch size of 4 and gradient accumulation of 8, giving an effective batch size of 32. Each stage trains for 1 epoch on up to 9,000 preference pairs, yielding approximately 281 optimiser steps. We use $\beta=0.3$.

For HelpSteer2 experiments, the attribute-specific datasets are smaller. We use a per-device batch size of 4 and gradient accumulation of 8, giving an effective batch size of 32. Each stage trains for 1 epoch, yielding approximately 58 steps for coherence and 69 steps for verbosity. We use $\beta=0.3$.

For PKU-SafeRLHF experiments, we use the same configuration as HH-RLHF, with per-device batch size of 4, gradient accumulation of 8, effective batch size of 32, 1 epoch, and $\beta=0.3$.

For UltraFeedback experiments, we use a per-device batch size of 2 and gradient accumulation of 4, giving an effective batch size of 8. Each stage trains for 1 epoch on 7,000 training pairs, yielding 875 optimiser steps. We use $\beta=0.3$.

\subsection{Sequential Reference Policy}
\label{app:reference_policy}

For every DPO stage, the reference policy is the frozen model state at the beginning of that stage. Stage~1 uses the initial base model as its reference, and Stage~2 uses the Stage~1 final adapter as its reference. This ensures each stage optimises relative to the model it starts from, matching practical sequential fine-tuning pipelines.

\subsection{Dataset Preprocessing}
\label{app:preprocessing}

All datasets are converted into the TRL conversational preference format with a user prompt, a chosen assistant response, and a rejected assistant response. The Llama-3.1 instruction chat template is applied using \texttt{apply\_chat\_template}. Prompt tokens are masked with label value $-100$ so that log-probability and loss computations are applied only to response tokens.

For evaluation, response log-probabilities are computed as mean per-token log-probabilities over non-masked response tokens. This length-normalised evaluation avoids systematically penalizing longer responses.

\subsection{Implementation}
\label{app:software}

Experiments are implemented using HuggingFace Transformers, TRL, PEFT, PyTorch, and Weights \& Biases. DPO training uses the TRL \texttt{DPOTrainer}. Adapter checkpoints are saved using PEFT after every stage. Training logs, gradient diagnostics, memory usage, and final metrics are recorded for each run. The codebase and preprocessing scripts will be released upon publication.

\begin{table*}[t]
\centering
\small
\resizebox{\textwidth}{!}{
\begin{tabular}{llccccc}
\toprule
Dataset & Conflict Type & Exact Label Conflict? & Previous Obj. Drop & Pairwise Pattern & Gradient Pattern & Main Interpretation \\
\midrule
HH-RLHF & Distributional & No & $-$0.218 / $-$0.503 & redistribution (easy pairs most affected) & near-orthogonal ($\theta{\approx}90^\circ$) & drift/rebalancing \\
HelpSteer2 & Multi-attribute & No exact flips; attribute-level & $-$0.273 / $-$0.205 & redistribution (easy pairs improve in vc; easy forget in cv) & near-orthogonal ($\theta{\approx}90^\circ$) & attribute transfer/drift \\
PKU-SafeRLHF & Extreme-signal safety & No exact flips; strong safety polarity & no margin drop (acc $-$9.9pp helpful) & easy pairs improve (hh); n.s. (hn) & near-orthogonal ($\theta{\approx}90^\circ$) & strong harmless signal persists \\
UltraFeedback & Compatible quality objectives & No & no drop; both improve & easy pairs improve most (both orders) & near-orthogonal ($\theta{\approx}90^\circ$) & positive transfer; no interference \\
\bottomrule
\end{tabular}
}
\caption{
Summary of conflict-type effects. The main claim is not that all sequential DPO causes the same forgetting pattern, but that the structure of the objective relationship changes both the behavioral and mechanistic signature of degradation.
}
\label{tab:conflict_taxonomy_results}
\end{table*}

\begin{table*}[t]
\centering
\small
\begin{tabular}{llccccc}
\toprule
Dataset & Order & Objective & Q1 Hardest & Q2 & Q3 & Q4 Easiest \\
\midrule
HH-RLHF & harmless$\rightarrow$helpful & Harmless & $+$0.025 & $+$0.024 & $-$0.005 & $+$0.009 \\
HH-RLHF & helpful$\rightarrow$harmless & Helpful  & $-$0.000 & $+$0.044 & $+$0.030 & $+$0.093 \\
HelpSteer2 & verbosity$\rightarrow$coherence & Verbosity & $+$0.011 & $-$0.005 & $-$0.016 & $-$0.083 \\
HelpSteer2 & coherence$\rightarrow$verbosity & Coherence & $-$0.035 & $-$0.007 & $+$0.007 & $+$0.019 \\
PKU-SafeRLHF & harmless$\rightarrow$helpful & Harmless & $-$0.032 & $-$0.055 & $-$0.084 & $-$0.151 \\
PKU-SafeRLHF & helpful$\rightarrow$harmless & Helpful  & $-$0.017 & $+$0.000 & $-$0.015 & $-$0.064 \\
UltraFeedback & IF$\rightarrow$honesty       & IF       & $-$0.106 & $-$0.104 & $-$0.240 & $-$0.805 \\
UltraFeedback & honesty$\rightarrow$IF       & Honesty  & $-$0.019 & $-$0.048 & $-$0.234 & $-$1.027 \\
\bottomrule
\end{tabular}
\caption{
Quartile analysis of pair-level margin drop. Pairs are sorted by Stage~1 margin and split into four equal-sized groups. Negative values indicate margin improvement; positive values indicate margin degradation.
}
\label{tab_app:quartile_results}
\end{table*}
\section{Mechanistic Diagnostic Implementation}
\label{app:mechanistic_implementation}

\subsection{Gradient Conflict Measurement Implementation}
\label{app:gradient_conflict}

Gradient conflict is measured during Stage~2 training to test whether the current training objective directly opposes the previous objective in parameter space. We implement this by subclassing \texttt{DPOTrainer} as \texttt{ConflictAwareDPOTrainer}, which overrides \texttt{training\_step}. After the standard DPO backward pass, the live Stage~2 gradients are available in the LoRA parameter gradients. We collect these gradients into a vector $\mathbf{g}_{\mathrm{current}}$.

We then temporarily clear gradients and compute a diagnostic DPO loss for the previous objective on a fixed reference batch. Backpropagating this diagnostic loss gives a second gradient vector $\mathbf{g}_{\mathrm{previous}}$. We compute cosine similarity between these two vectors:

\begin{equation}
\cos(\mathbf{g}_{\mathrm{current}}, \mathbf{g}_{\mathrm{previous}})
=
\frac{
\mathbf{g}_{\mathrm{current}} \cdot \mathbf{g}_{\mathrm{previous}}
}{
\|\mathbf{g}_{\mathrm{current}}\| \|\mathbf{g}_{\mathrm{previous}}\|
}.
\end{equation}

After measurement, the original Stage~2 gradients are restored so the optimiser step remains unchanged. All gradient vectors are moved to CPU immediately after collection to reduce GPU memory pressure.

Reference log-probabilities for the diagnostic previous-objective DPO loss are precomputed before training using the frozen stage-initial reference model. This avoids keeping an additional full reference model active during training. Measurements are taken at fixed intervals, typically every 100 optimiser steps. The diagnostic batch size is selected based on memory constraints and reported with the corresponding experiment.

\subsection{Gradient Conflict Interpretation}
\label{app:gradient_interpretation}

The cosine similarity is interpreted as follows:

\begin{itemize}
    \item A cosine near $+1$ means the gradients are aligned. Improving the current objective also improves the previous one.
    \item A cosine near $0$ means the gradients are approximately orthogonal. The objectives are not directly opposed in LoRA gradient space.
    \item A cosine near $-1$ means the gradients are opposed. Improving the current objective directly harms the previous one.
\end{itemize}

Because gradients are measured only over trainable LoRA parameters and on a fixed diagnostic batch, this analysis should be interpreted as a LoRA-space diagnostic rather than as a full-model causal proof.

\subsection{Adapter Movement Analysis}
\label{app:adapter_movement}

As an auxiliary diagnostic, we analyse movement between LoRA adapter checkpoints. In the primary reported analysis, we flatten and concatenate all trainable LoRA $A$ and $B$ matrices into a raw adapter vector for each stage. We then compute L2 distances and cosine similarities between stage vectors and update directions.

For stages $S_1$ and $S_2$, we define the raw update direction:

\begin{equation}
\delta_{1\rightarrow2} = S_2 - S_1
\end{equation}

We compute:

\begin{itemize}
    \item $\|S_1\|$ is the magnitude of the Stage~1 adapter.
    \item $\|S_2-S_1\|$ is the movement from Stage~1 to Stage~2.
    \item $\cos(S_1, S_2-S_1)$ indicates whether the Stage~2 update moves with, against, or orthogonal to the Stage~1 adapter.
\end{itemize}

We treat this analysis as supporting evidence because raw LoRA parameter space is not invariant to LoRA rescaling. Effective LoRA updates of the form $BA$ provide a more direct representation of the induced weight change, but dense materialization of all effective updates can be memory intensive for 8B models. Therefore, raw adapter movement is reported only as an auxiliary parameter-space diagnostic.

\section{Cross-Setting Summary}
\label{app:setting_summary}

Across the four settings, sequential DPO produces different behavioural signatures depending on the relationship between objectives. HH-RLHF shows partial degradation under a distributional safety--helpfulness shift; HelpSteer2 shows attribute-level redistribution; PKU-SafeRLHF shows stability of the stronger harmlessness objective; and UltraFeedback shows positive transfer between compatible response-quality objectives. Table~\ref{tab:conflict_taxonomy_results} summarizes these patterns alongside the corresponding pair-level and mechanistic diagnostics.

\section{Dataset and Preference-Relationship Details}
\label{app:dataset_details}

This appendix provides additional details on the four preference settings used in the main experiments. Each setting consists of two objectives, and we train both objective orderings under the same sequential DPO and cross-evaluation protocol.

\paragraph{HH-RLHF: Distributional conflict.}
Anthropic HH-RLHF contains separate helpfulness and harmlessness preference subsets~\cite{rlhf_paper}. In our setup, we use \emph{helpful-base} and \emph{harmless-base} to instantiate a distributional safety--helpfulness shift. The \emph{harmless-base} subset contains safety-oriented chosen--rejected comparisons, where the preferred response is the safer or less harmful response, while the \emph{helpful-base} subset contains assistant-response comparisons oriented toward helpfulness. The two subsets differ in both prompt distribution and annotated preference dimension. Harmlessness targets safety-relevant behaviour, whereas helpfulness targets general assistant quality.

We use these splits to introduce a distributional conflict rather than an adversarial label-conflict setting. Empirically, we find no exact flipped-label pairs across our training sets and less than $1\%$ prompt overlap under exact string matching ($0.86\%$), suggesting that the conflict is primarily distributional. For HH-RLHF, each objective uses up to 9,000 training pairs and 1,000 held-out test pairs.

\paragraph{HelpSteer2: Multi-attribute interaction.}
HelpSteer2~\cite{helpsteer2} provides scalar ratings across response attributes including helpfulness, correctness, coherence, complexity, and verbosity, each on a 0--4 Likert scale. We use this dataset to construct preference objectives for verbosity and coherence. Verbosity reflects the amount of detail or elaboration, while coherence reflects whether a response is logically organised and easy to follow. Because increased detail can sometimes introduce repetition, tangents, or loss of focus, verbosity and coherence provide a controlled attribute-level setting distinct from broad safety--helpfulness comparisons.

For each attribute, we construct within-prompt preference pairs by selecting response pairs whose target-attribute rating differs by at least one point. Cross-prompt pairing is excluded to avoid confounding response quality with prompt difficulty. After filtering and deduplication, we obtain 2,229 verbosity training pairs and 1,878 coherence training pairs. The held-out test sets contain 527 verbosity pairs and 422 coherence pairs. We run both verbosity$\rightarrow$coherence and coherence$\rightarrow$verbosity orderings.

\paragraph{PKU-SafeRLHF: Strong safety signal.}
PKU-SafeRLHF~\cite{safe_rlhf_pku} provides response pairs annotated for both harmlessness and helpfulness. We use it to study a setting where harmlessness provides a much stronger observed preference signal than in HH-RLHF. In our construction, many harmlessness pairs contrast safer refusal-style responses with clearly unsafe responses. This makes the dataset useful for testing whether a high-signal safety objective is more resistant to forgetting during later helpfulness training.

We construct harmless and helpful preference splits and run both harmless$\rightarrow$helpful and helpful$\rightarrow$harmless orderings. We use the provided helpful split without additional filtering because the chosen helpful responses are predominantly benign in our construction. For PKU-SafeRLHF, each objective uses up to 9,000 training pairs and 1,000 held-out test pairs.

\paragraph{UltraFeedback: Compatible objectives.}
UltraFeedback~\cite{cui2023ultrafeedback} provides LLM-generated responses with fine-grained scalar annotations across multiple response-quality dimensions, including instruction-following, honesty, truthfulness, and helpfulness. We use instruction-following and honesty to study compatible objectives. Both objectives reward high-quality responses that address the prompt accurately and faithfully, making them conceptually aligned rather than directly conflicting.

For each attribute, we construct 7,000 training pairs and 700 held-out test pairs. We run both instruction-following$\rightarrow$honesty and honesty$\rightarrow$instruction-following orderings. This setting serves as a compatible-objective baseline. When two objectives reward overlapping response behaviours, sequential training should produce stability or positive transfer rather than interference.

\end{document}